\documentclass[a4paper,conference]{IEEEtran}

\usepackage{cite}

\usepackage[pdftex]{graphicx}
\usepackage{amsmath}
\usepackage{amssymb}
\usepackage{algorithmic}
\usepackage{array}
\usepackage{mathtools}

\usepackage{multirow}

\usepackage{etoolbox}
\makeatletter
\patchcmd{\@makecaption}
  {\scshape}
  {}
  {}
  {}
\makeatother

\ifCLASSOPTIONcompsoc
  \usepackage[caption=false,font=normalsize,labelfont=sf,textfont=sf]{subfig}
\else
  \usepackage[caption=false,font=footnotesize]{subfig}
\fi
\usepackage{epstopdf}
\usepackage{float}

\usepackage{fixltx2e}

\usepackage{stfloats}
\usepackage{url}

\usepackage[dvipsnames]{xcolor}
\newcommand{\SV}[1]{#1}
\newcommand{\AT}[1]{#1}

\begin{document}

\title{Transductive Label Augmentation for Improved Deep Network Learning}

\author{\IEEEauthorblockN{Ismail Elezi\footnotemark{*}}
\IEEEauthorblockA{University of Venice, Italy\\
				  ZHAW Datalab, Switzerland \\
ismail.elezi@unive.it}
\and
\IEEEauthorblockN{Alessandro Torcinovich\footnotemark{*}}
\IEEEauthorblockA{University of Venice, Italy\\ 
DAIS\\
ale.torcinovich@unive.it }
\and
\IEEEauthorblockN{Sebastiano Vascon\footnotemark{*}}
\IEEEauthorblockA{University of Venice, Italy\\
				  DAIS/ECLT\\
sebastiano.vascon@unive.it}
\and
\IEEEauthorblockN{Marcello Pelillo}
\IEEEauthorblockA{University of Venice, Italy\\
				  DAIS/ECLT\\
pelillo@unive.it}}

\maketitle

\begin{abstract}
A major impediment to the application of deep learning to real-world problems is the scarcity of labeled data. Small training sets are in fact of no use to deep networks as, due to the large number of trainable parameters, they will very likely be subject to overfitting phenomena.
On the other hand, the increment of the training set size through further manual or semi-automatic labellings can be costly, if not possible at times. Thus, the standard techniques to address this issue are transfer learning and data augmentation, which consists of applying some sort of ``transformation'' to existing labeled instances to let the training set grow in size. Although this approach works well in applications such as image classification, where it is relatively simple to design suitable transformation operators, it is not obvious how to apply it in more structured scenarios. Motivated by the observation that in virtually all application domains it is easy to obtain unlabeled data, in this paper we take a different perspective and propose a \emph{label augmentation} approach. We start from a small, curated labeled dataset and let the labels propagate through a larger set of unlabeled data using graph transduction techniques. This allows us to naturally use (second-order) similarity information which resides in the data, a source of information which is typically neglected by standard augmentation techniques. In particular, we show that by using known game theoretic transductive processes we can create larger and accurate enough labeled datasets which use results in better trained neural networks. Preliminary experiments are reported which demonstrate a consistent improvement over standard image classification datasets.
\end{abstract}

\makeatletter{\renewcommand*{\@makefnmark}{}
\footnotetext{* = Equal contribution. Authors are listed in alphabetical order.}\makeatother}

\IEEEpeerreviewmaketitle

\section{Introduction}
Deep neural networks (DNNs) have met with success multiple tasks, and testified a constantly increasing popularity, being able to deal with the vast heterogeneity of data and to provide state-of-the-art results across many fields and domains \cite{DBLP:journals/nature/LeCunBH15,DBLP:journals/nn/Schmidhuber15}.
Convolutional Neural Networks (CNNs) \cite{DBLP:journals/pr/FukushimaM82,DBLP:journals/neco/LeCunBDHHHJ89} are one of the protagonists of this success. Starting from AlexNet \cite{DBLP:conf/nips/KrizhevskySH12}, until the most recent convolutional-based architectures \cite{DBLP:conf/cvpr/SzegedyLJSRAEVR15,DBLP:conf/cvpr/HeZRS16,DBLP:conf/cvpr/HuangLMW17} CNNs have proved to be especially useful in the field of computer vision, improving the classification accuracy in many datasets \cite{DBLP:conf/cvpr/DengDSLL009,Krizhevsky09learningmultiple}.

However, a common caveat of large CNNs is that they require a lot of training data in order to work well. In the presence of classification tasks on small datasets, typically those networks are \emph{pre-trained} in a very large dataset like ImageNet \cite{DBLP:conf/cvpr/DengDSLL009}, and then \emph{finetuned} on the dataset the problem is set on. The idea is that the pre-trained network has stored a decent amount of information regarding features which are common to the majority of images, and in many cases this knowledge can be transferred to different datasets or to solve different problems (image segmentation, localization, detection, etc.). This technique is referred as \emph{transfer learning} \cite{DBLP:conf/nips/YosinskiCBL14} and has been an important ingredient in the success and popularization of CNNs. Another important technique -- very often paired with the previous one -- is \emph{data augmentation}, through which small transformations are directly applied on the images. A nice characteristic of data augmentation is its agnosticism toward algorithms and datasets. \cite{DBLP:conf/cvpr/CiresanMS12} used this technique to achieve state-of-the-art results in MNIST dataset \cite{lecun-mnisthandwrittendigit-2010}, while \cite{DBLP:conf/nips/KrizhevskySH12} used the method almost without any changes to improve the accuracy of their CNN in the ImageNet dataset \cite{DBLP:conf/cvpr/DengDSLL009}. Since then, data augmentation has been used in virtually every implementation of CNNs in the field of computer vision.
% AT: in my opinion this last sentence does not explain the adaptability toward algorithms and datasets

Despite the practicality of the above-mentioned techniques, when the number of images per class is extremely small, the performances of CNNs rapidly degrade and leave much to be desired. The high availability of unlabeled data only solves  half of the problem, since the manual labeling process is usually costly, tedious and prone to human error. Under these assumptions, we propose a new method to perform an automatic labeling, called \emph{transductive label augmentation}. Starting from a very small labeled dataset, we set an automatic label propagation procedure, that relies on graph transduction techniques, to label a large unlabeled set of data. This method takes advantage of second-order similarity information among the data objects, a source of information which is not directly exploited by traditional techniques. To assess our statements, we perform a series of experiments with different CNN architectures and datasets, comparing the results with a first-order ``label propagator''.

In summary, our contributions in this article are as follows: a) by using graph transductive approaches, we propose and develop the aforementioned label augmentation method and use it to improve the accuracy of state-of-the-art CNNs in datasets where the number of labels is limited; b) by gradually increasing the number of labeled objects, we give detailed results in three standard computer vision datasets and compare the results with the results of CNNs; c) we replace our transductive algorithm with linear support vector machines (SVM) \cite{DBLP:journals/ml/CortesV95} to perform label augmentation and compare the results; d) we give directions for future work and how the method can be used on other domains. 

\begin{figure*}[ht!]
 \centering
 \includegraphics[scale=0.40,trim={1cm 2.4cm 0cm 0cm},clip]{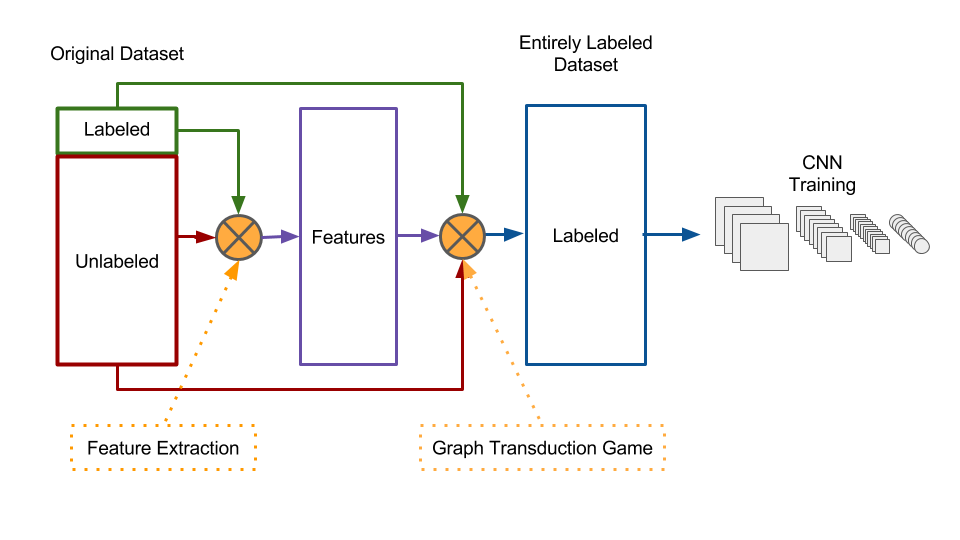}
 \caption{The pipeline of our method. The dataset consists of labeled and unlabeled images. First, we extract features from the images, and then we feed the features (and the labels of the labeled images) to graph transduction games. For the unlabeled images, we use a uniform probability distribution as 'soft-labeling'. The final result is that the unlabeled points get labeled, thus the entire dataset can be used to train a convolutional neural network.}
\end{figure*}

\subsection{Related Work}

Semi-supervised label propagation has a long history of usage in the field of machine learning \cite{vapnik}. Starting from an initial large dataset, with a small portion of labeled observations the traditional way of using semi-supervised learning is to train a classifier only in the labeled part, and then use the classifier to predict labels for the unlabeled part. The labels predicted in this way are called \emph{pseudo-labels}. The classifier is then trained in the entire dataset, considering the pseudo-labels as if they were real labels.

Different methods with the same intent have been previously proposed. In deep learning in particular, there have been devised algorithms to use data with a small number of labeled observations. \cite{Lee_pseudo-label:the} trained the network jointly in both the labeled and unlabeled points. The final loss function is a weighted loss of both labeled and unlabeled points, where in the case of the unlabeled points, the pseudo-label is determined by the highest score proposed by the model. \cite{DBLP:conf/cvpr/HausserMC17} optimized a CNN on such a way as to produce embeddings that have high similarities for the observations that belong to the same class. \cite{DBLP:conf/nips/KingmaMRW14} used a totally different approach, developing a generative model that allows for effective generalization from small labeled datasets to large unlabeled ones.

In all the mentioned methods, the way how the unlabeled data has been used can be considered as an intrinsic property of their engineered neural networks. Our choice of CNNs as the algorithm used for the experiments was motivated because CNNs are state-of-the-art models in computer vision, but the approach is more general than that. The method presented in this article does not even require a neural network and in principle, non-feature based observations (i.e graphs) can be considered, as long as a similarity measure can be derived for them. At the same time, the method shows good results in relatively complex image datasets, improving over the results of state-of-the-art CNNs.

\section{Graph Transduction Game}

Graph Transduction (GT) is a subfamily of semi-supervised learning that aims to classify unlabeled objects starting from a small set of labeled ones. In particular, in GT the data is modeled as a graph whose vertices are the objects in a dataset. The provided label information is then propagated all over the unlabeled objects through the edges, weighted according to the consistency of object pairs. The reader is encouraged to refer to \cite{ZhuSemiSupervised} for a detailed description of algorithms and applications on graph transduction.

\begin{figure}[t!]
 \centering
 \includegraphics[width=\linewidth,trim={2.8cm 0.5cm 2.4cm 0.5cm},clip]{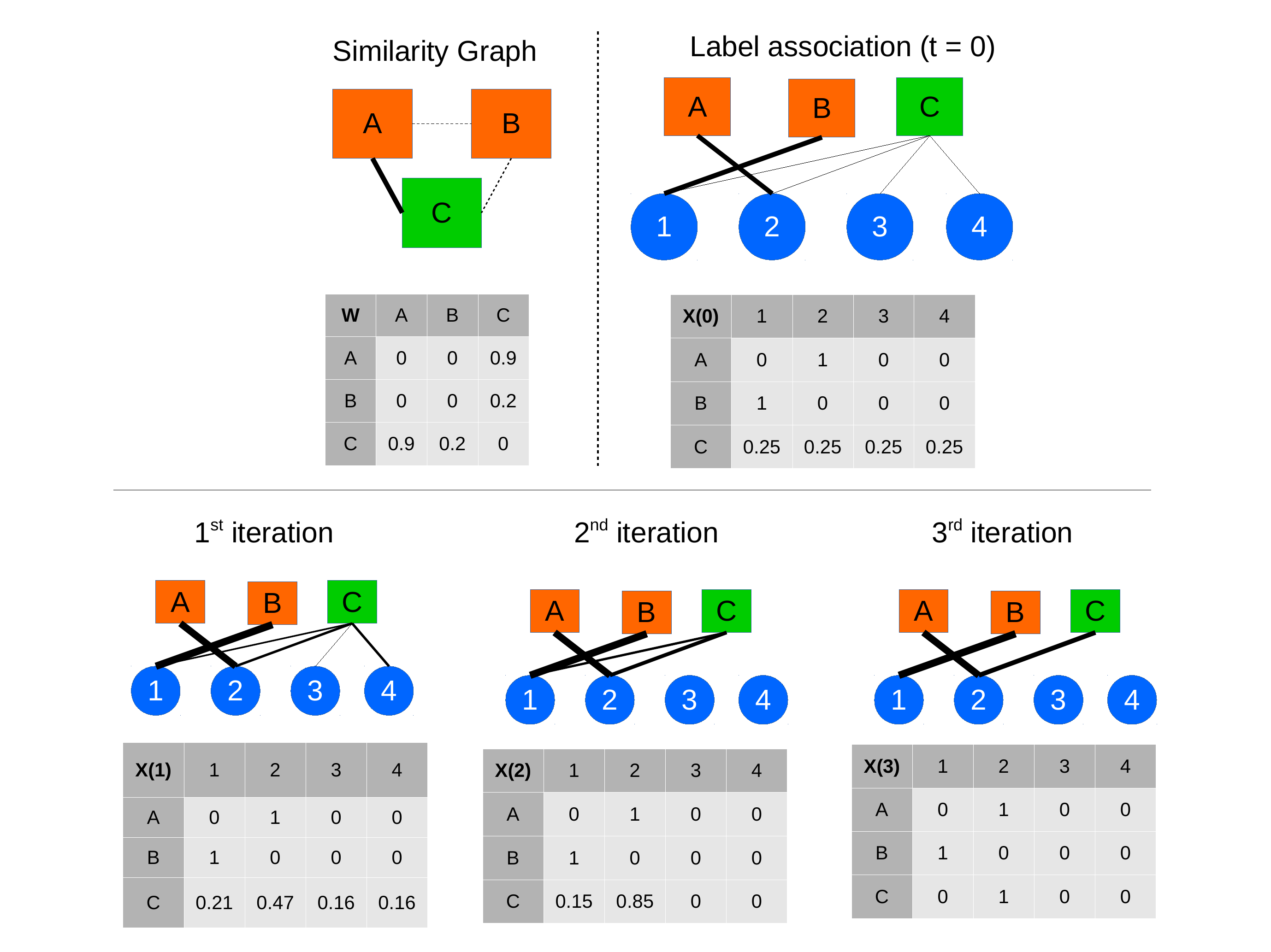}
%\vspace{-0.5cm}
 \caption{The dynamics of the GTG. The algorithm takes in input similarities between objects and hard/soft labelings of the object themselves. After three iterations, the algorithm has converged, generating a pseudo-label with 100\% confidence.}
%\vspace{-0.5cm}
\end{figure}

More formally, let $G = (V, E, w)$ be a graph. $V$ is the vertex set of the objects and can be partitioned in two sets: $L = \{(f_1, y_1), ..., (f_l, y_l)\}$ contains the labeled objects, where $f_i \in \mathbb{R}^d$ is a real-valued vector describing the object (features), and $y_i \in \{1, 2...,m\}$ is its associated label, while $U = \{f_{l+1},...,f_{n}\}$ is the set of unlabeled objects. $E$ is the set of edges connecting the vertices and $w: E \rightarrow \mathbb{R}_{\geq 0}$ is a weight function that assigns a non-negative similarity measure to each edge in $E$, and can be summarized in a weight matrix $W$.

In \cite{ZhuSemiSupervised}, GT takes in input $W$ along with initial probability distributions for every objects -- one-hot labels for $(f_i, y_i) \in L$, soft labels for $f_i \in U$ -- and iteratively applies a function $P: \Delta^m \rightarrow \Delta^m$ where $\Delta^m$ is the standard simplex. At each iteration, if the distributions of labeled objects have changed, they are reset. Once the algorithm reaches the convergence, the resulting final probabilities give a labeling over the entire set of objects.

In this article, we follow the approach proposed in \cite{DBLP:journals/neco/ErdemP12}, where the authors interpret the graph transduction task as a non-cooperative multiplayer game. The same methodology has been successfully applied in different context, e.g. bioinformatics \cite{DBLP:journals/jmiv/vascon2018} and matrix factorization \cite{DBLP:conf/icpr/TripodiVP16}.

%\input{gtg_seba_new}
%%%%%%%%%%%%%%%%%%%%%%%%%%%%
\SV{In graph transduction game (GTG), 
% we represent the objects of a dataset as players and their possible labels as strategies.
\AT{objects of a dataset are represented as players and their labels as strategies.}
% AT: I have use passive form for all the theory so I would like to keep it coherent
\AT{In synthesis,} % Then
a non-cooperative multiplayer game is played among the objects, until an equilibrium condition is reached, the \emph{Nash Equilibria} \cite{Nash1951}. 
Here, we provide some basic knowledge on game theory in order to be self-contained. Given a set of players $I = \{1, \dots, n\}$ and a set of possible \AT{pure} strategies $S = \{1, \dots, m\}$:
\begin{enumerate}
\item \emph{mixed strategy}: a mixed strategy $x_i$ is a probability distribution over the possible strategies for player $i$. Then $x_i \in \Delta^m$, where 
$$\Delta^m = \left\lbrace \sum_{h = 1}^{m} x_i(h) = 1, x_i(h) \geq 0, \ h = \{1, \dots, m\} \right\rbrace $$
% AT: I would take out this definition, simply stating the it's the m-dimensional standard simplex
is the standard $m$-dimensional simplex and $x_i(h)$ is the probability of player $i$ choosing \AT{the pure} strategy $h$.
% Each player $i$ has an associated mixed strategy which encapsulates the probability of picking a certain strategy. 
\item \emph{\AT{mixed} strategy space}: \AT{it} corresponds to the set of all  mixed strategies of the players
%in the game 
$x = \left\lbrace x_1, \dots, x_n \right\rbrace$
% \item \emph{utility function}: 
%is a function that
% \AT{it} maps mixed strategies to payoffs\AT{. In particular} $u: \Delta^m \rightarrow \mathbb{R}_{\geq 0}$ and represents the gain obtained by a player when it plays a certain strategy.
\item \emph{utility function}: it represents the gain obtained by a player when it chooses a certain mixed strategy, in particular $u: \Delta^m \rightarrow \mathbb{R}_{\geq 0}$.
\end{enumerate}
%Here we assume
\AT{Here, it is assumed} that the payoffs associated to each player are additively separable, 
\AT{thus the algorithm is a member of \AT{polymatrix games}\cite{howson1972equilibria}.}
%making the proposed non-cooperative game a member of a class of games known as polymatrix games\cite{8_7}.
In GTG, the \AT{aforementioned definitions}
% definitions introduced above
turns into the following:}
%AT: This part can be summarized and put in another section (I would personally cite Strategy space), I already explained how to initialize starting probabilities and I it is already written 3 times in the paper
\SV{\paragraph{Strategy space} The strategy space $x$ is the starting point of the game and contains all the mixed strategies. The space $x$ can be initialized in different ways based on the fact that some prior knowledge exists or not. Here, we distinguish the initialization based on the type of object, \emph{labeled} or \emph{unlabeled}.
For the labeled object, since their class is known, a one-hot vector is assigned:
\begin{equation}\label{eq:Xprior}
x_i(h)=
\begin{cases}
	1, & \text{if } i \text{ has label } h \\
    0, & \text{ otherwise}.
\end{cases}
\end{equation}. 
For the unlabeled objects all the labels have the same probability of being associated to an object, thus:
\begin{equation}
x_i(h) = \frac{1}{m} \quad \text{ } h = \{1, \dots, m \}\label{eq:Xnoprior}
\end{equation}}

%AT this function speaks about the utility function, NOT the payoff one! It should be merged with the previous paragraph for the payoff function
\SV{\paragraph{Payoff function} 
The utility function reflects the likelihood of choosing a particular label and considers the similarity between labeled and unlabeled players. Similar players influence each other more in picking one of the possible strategies (labels). Once the game reaches an equilibrium, every player play their best strategies  which correspond to a consistent labeling \cite{miller1991copositive} not only for the player itself but also for the others.
Under equilibrium conditions the label of player $i$ is given by the strategy played with the highest probability. %The ingredients for the GTG are the following: a set of \emph{players}, a \emph{payoff function} that quantifies the gains for each player and a method for finding a \emph{Nash Equilibria}.
Formally, given a player $i$ and a strategy $h$:
\begin{eqnarray}
u_i(h)=\sum_{j \in U}{(A_{ij}x_j)_h}+\sum_{k=1}^{m}{\sum_{j \in L}{A_{ij}(h,k)}} \\
u_i(x)=\sum_{j \in U}{x_i^TA_{ij}x_j}+\sum_{k=1}^{m}{\sum_{j \in L}{x_i^T(A_{ij})_k}}
\end{eqnarray}
where $u_i(x)$ is the utility received by player $i$ when it plays the mixed strategy $x_i$ and $A_{ij} \in \mathbb{R}^{m \times m}$ is the \emph{partial payoff matrix} between players $i$ and $j$. As in \cite{DBLP:journals/neco/ErdemP12}, 
%the partial payoff
$A_{ij} = I_m \times \omega_{ij}$ where $\omega_{ij}$ is the similarity between player $i$ and $j$ and $I_m$ is the identity matrix of size $m \times m$. %The rationale is that the higher the similarity between pair of players the more they will affect each other in the final labeling.\\
% The following should go in experimental part
The similarity function between players (objects) can be given or computed starting from the features. Given two objects $i, j$ and their features $f_i$, $f_j$, their similarity is computed following the method proposed by \cite{DBLP:conf/nips/Zelnik-ManorP04}:
\begin{equation}\label{eq:payoff}
\omega(i,j)=exp\left\lbrace-\frac{||f_i-f_j||_2}{\sigma_{i}\text{ }\sigma_{j}}\right\rbrace
\end{equation}
where $\sigma_{i}$ corresponds to the distance between $i$ and its $7$-nearest- neighbors. Similarity values are stored in matrix $W$.

\paragraph{Finding Nash Equilibria}
The last component of our method is an algorithm for finding equilibrium conditions in this game. In \cite{DBLP:journals/neco/ErdemP12} a result from Evolutionary Game Theory \cite{weibull1997evolutionary}, named Replicator Dynamics (RD) \cite{smith1982evolution} is used. The RD are a class of dynamical systems that perform a natural selection process on a multi-population of strategies. The idea is to lead the fittest strategies to survive while the others to go extinct. More specifically the RD are defined as follow: 
\begin{equation}
x_i(h)^{t + 1} = x_i(h)^{t}\frac{u_i(h)^t}{u_i(x^t)}
\end{equation}
where $x_i(h)^{t}$ is the probability of strategy $h$ at time $t$ for player $i$.\\
The RD are iterated until convergence, this means either the distance between two successive steps is zero (formally $||x^{t+1}-x^t||_2 \leq \varepsilon$) or a certain amount of iterations is reached (See \cite{DBLP:journals/jmiv/Pelillo97} for a detailed analysis). In practical applications one could set the $\varepsilon$ to a small number \AT{but typically} 10-20 iterations are sufficient.}
%while for the maximum number of iterations typically 10-20 iterations are sufficient.}
%%%%%%%%%%%%%%%%%%%%%%%%%%%%%

\section{Label Generation}
The previously explained framework can be applied to a dataset with many unlabeled objects to perform an automatic labeling and thus increase the availability of training objects. In this article we deal with datasets for image classification, but our approach can be applied in other domains too.

\textbf{Preliminary step}: both the labeled and unlabeled sets can be refined to obtain more informative feature vectors. In this article, we used fc7 features of CNNs trained on ImageNet, but in principle, any type of features can be considered. Our particular choice was motivated because fc7 features work significantly better than traditional computer vision features (SIFT \cite{DBLP:journals/ijcv/Lowe04} and its variations). While this might seem counter-intuitive (using pre-trained CNNs on ImageNet, while we are solving the problem of limited labeled data), we need to consider that our datasets are different from ImageNet (they come from different distributions), and by using some other dataset to pre-train our networks, we are not going against the spirit of the idea of the paper.

\textbf{Step 1}: the objects are assigned to initial probability distributions, needed to start the GTG. The labeled ones use their respective one-hot label representations, while the unlabeled ones can be set to a uniform distribution among all the labels. In presence of previous possessed information, some labels can be directly excluded in order to start from a multi-peaked distribution, which if chosen wisely, can improve the final results.

\textbf{Step 2}: the extracted features are used to compute the similarity matrix $W$. The literature \cite{DBLP:conf/nips/Zelnik-ManorP04} presents multiple methods to obtain a $W$ matrix 
and extra care should be taken when performing this step, since an incorrect choice in its computation can determine a failure in the transductive labeling.

\textbf{Step 3}: once $W$ is computed, graph transduction game can be played (up to convergence) among the objects to obtain the final probabilities which determine the label for the unlabeled objects.

The resulting labeled dataset can then be used to train a classification model. This is very convenient for several reasons: 1) CNNs are fully parametric models, so we do not need to store the training set in memory like in the case of graph transduction. In some aspect, the CNN is approximating in a parametric way the GTG algorithm; 2) the inference stage on CNNs is extremely fast (real-time); 3) CNN features can be used for other problems, like image segmentation, detection and classification, something that we cannot do with graph-transduction or with classical machine learning methods (like SVM). In the next section we will report the results obtained from state-of-the-art CNNs, and compare those results with the same CNNs trained only on the labeled part of the dataset.

\section{Experiments}
%\input{tab_res}
%%%%%%%%%%%%%%%%%%%%%%%%%%%
\begin{table}[h!]
\vspace{0.3cm}
\label{tab:results}
\footnotesize
\centering
\begin{tabular}{|c|c|c|c|c|c|c|}
\hline
\multirow{2}{*}{\begin{tabular}[c]{@{}c@{}}accuracy\\ 2\% labeled \end{tabular}} & \multicolumn{2}{c|}{caltech}                       & \multicolumn{2}{c|}{indoors}                       & \multicolumn{2}{c|}{scenenet}                      \\ \cline{2-7} 
                                                                        & RN18 & \multicolumn{1}{c|}{DN121} & RN18                  & \multicolumn{1}{c|}{DN121} & RN18                  & \multicolumn{1}{c|}{DN121} \\ \hline
		   GTG + CNN & \textbf{0.532} & \textbf{0.620} & \textbf{0.486} & \textbf{0.538} & \textbf{0.430} &	\textbf{0.495} \\
		   SVM + CNN & 			0.473 & 	     0.539 &      	  0.434 & 		   0.468 &		  	0.370 & 		 0.417 \\
                 CNN &			0.266 & 		 0.235 & 		  0.341 & 		0.323 & 			0.205 & 		 0.178 \\
\hline
\hline
\multirow{2}{*}{\begin{tabular}[c]{@{}c@{}}F score\\ 2\% labeled\end{tabular}} & \multicolumn{2}{c|}{caltech}                       & \multicolumn{2}{c|}{indoors}                       & \multicolumn{2}{c|}{scenenet}                      \\ \cline{2-7} 
                                                                        & RN18                  & \multicolumn{1}{c|}{DN121} & RN18                  & \multicolumn{1}{c|}{DN121} & RN18                  & \multicolumn{1}{c|}{DN121} \\ \hline
		   GTG + CNN & \textbf{0.468} & \textbf{0.559} & \textbf{0.357} & \textbf{0.396} & \textbf{0.399} &	\textbf{0.457} \\
		   SVM + CNN & 			0.388 & 	     0.455 &      	  0.319 & 		   0.327 &		  	0.352 & 		 0.377 \\
                 CNN &			0.181 & 		 0.151 & 		  0.187 & 		   0.172 & 			0.191 & 		 0.167 \\
\hline
\end{tabular}
\vspace{1.5mm}
\caption{The results of our algorithm, compared with the results of linear SVM and CNN, when only 2\% of the dataset is labeled. We see that in all three datasets and two different neural networks, our approach gives significantly better results than SVM or CNN.} 
\end{table}

\begin{table}[t!]
\label{tab:results}
\footnotesize
\centering
\begin{tabular}{|c|c|c|c|c|c|c|}
\hline
\multirow{2}{*}{\begin{tabular}[c]{@{}c@{}}accuracy\\ 5\% labeled\end{tabular}} & \multicolumn{2}{c|}{caltech}                       & \multicolumn{2}{c|}{indoors}                       & \multicolumn{2}{c|}{scenenet}                      \\ \cline{2-7} 
                                                                        & RN18                  & \multicolumn{1}{c|}{DN121} & RN18                  & \multicolumn{1}{c|}{DN121} & RN18                  & \multicolumn{1}{c|}{DN121} \\ \hline
		   GTG + CNN & \textbf{0.625} & \textbf{0.698} & \textbf{0.568} & \textbf{0.613} & \textbf{0.563} &	\textbf{0.621} \\
		   SVM + CNN & 			0.605 & 	     0.675 &      	  0.516 & 		   0.580 &		  	0.511 & 		 0.601 \\
	             CNN &			0.457 & 		 0.444 & 		  0.456 & 		0.466 & 			0.408 & 		 0.438 \\
\hline
\hline
\multirow{2}{*}{\begin{tabular}[c]{@{}c@{}}F score\\ 5\% labeled\end{tabular}} & \multicolumn{2}{c|}{caltech}                       & \multicolumn{2}{c|}{indoors}                       & \multicolumn{2}{c|}{scenenet}                      \\ \cline{2-7} 
                                                                        & RN18                  & \multicolumn{1}{c|}{DN121} & RN18                  & \multicolumn{1}{c|}{DN121} & RN18                  & \multicolumn{1}{c|}{DN121} \\ \hline
		   GTG + CNN & \textbf{0.571} & \textbf{0.653} & \textbf{0.454} & \textbf{0.508} & \textbf{0.536} &	\textbf{0.608} \\
		   SVM + CNN & 			0.542 & 	     0.626 &      	  0.426 & 		   0.505 &		  	0.501 & 		 0.590 \\
                 CNN &			0.372 & 		 0.358 & 		  0.345 & 		   0.306 & 			0.394 & 		 0.419 \\
\hline
\end{tabular}
\vspace{1.5mm}
\caption{The results of our algorithm, compared with the results of linear SVM and CNN, when 5\% of the dataset is labeled.} 
\end{table}

\begin{table}[t!]
\label{tab:results}
\footnotesize
\centering
\begin{tabular}{|c|c|c|c|c|c|c|}
\hline
\multirow{2}{*}{\begin{tabular}[c]{@{}c@{}}accuracy \\ 10\% labeled\end{tabular}} & \multicolumn{2}{c|}{caltech}                       & \multicolumn{2}{c|}{indoors}                       & \multicolumn{2}{c|}{scenenet}                      \\ \cline{2-7} 
                                                                        & RN18                  & \multicolumn{1}{c|}{DN121} & RN18                  & \multicolumn{1}{c|}{DN121} & RN18                  & \multicolumn{1}{c|}{DN121} \\ \hline
		   GTG + CNN & \textbf{0.667} & \textbf{0.727} & \textbf{0.598} & \textbf{0.645} & \textbf{0.624} &	\textbf{0.686} \\
		   SVM + CNN & 			0.658 & 	     0.724 &      	  0.576 & 		   0.635 &		  	0.622 & 		 0.660 \\
           		 CNN &			0.577 & 		 0.598 & 		  0.553 & 		   0.567 & 			0.571 & 		 0.584 \\
\hline
\hline
\multirow{2}{*}{\begin{tabular}[c]{@{}c@{}}F score \\ 10\% labeled\end{tabular}} & \multicolumn{2}{c|}{caltech}                       & \multicolumn{2}{c|}{indoors}                       & \multicolumn{2}{c|}{scenenet}                      \\ \cline{2-7} 
                                                                        & RN18                  & \multicolumn{1}{c|}{DN121} & RN18                  & \multicolumn{1}{c|}{DN121} & RN18                  & \multicolumn{1}{c|}{DN121} \\ \hline
		   GTG + CNN & \textbf{0.622} & \textbf{0.694} & 		  0.509 & 		   0.574 & 			0.609 &	\textbf{0.700} \\
		   SVM + CNN & 			0.612 & 	     0.686 & \textbf{0.515} & \textbf{0.579} & \textbf{0.612} & 		 0.650 \\
                 CNN &			0.519 & 		 0.533 & 		  0.478 & 		   0.471 & 			0.565 & 		 0.570 \\
\hline
\end{tabular}
\vspace{1.5mm}
\caption{The results of our algorithm, compared with the results of linear SVM and CNN, when  10\% of the dataset is labeled.} 
\vspace{-0.5cm}
\end{table}

% baseline & 			0.25211 & 		   0.29783 & 		  0.34153 & 		 0.31143 & 			0.23133 & 			 0.207 \\
% SVM 	 & 			0.46573 & 		   0.54657 & 		  0.44944 & 		 0.48887 & 			0.38567 & 			 0.427 \\
% GTG 	 & \textbf{0.52425} & \textbf{0.59301} & \textbf{0.50541} & \textbf{0.50965} & \textbf{0.47533} & \textbf{0.48767} \\
%%%%%%%%%%%%%%%%%%%%%%%%%%%%%%%%%%%%%%

In order to assess the quality of the algorithm, we used it to automatically label three known realistic datasets, namely \emph{Caltech-256} \cite{caltech}, \emph{Indoor Scene Recognition} \cite{DBLP:conf/cvpr/QuattoniT09} and \emph{SceneNet-100} \cite{DBLP:conf/eccv/KadarB14}. \emph{Caltech-256} contains $30607$ images belonging to $256$ different categories and it is used for object recognition tasks. \emph{Indoor Scene Recognition} is a dataset containing $15620$ images of different common places (restaurants, bedrooms, etc.), divided in $67$ categories and, as the name says, it is used for scene recognition. \emph{SceneNet-100} database is a publicly available online ontology for scene understanding that organizes scene categories according to their perceptual relationships. The dataset contains $10000$ real-world images, separated into $100$ different classes.

Each dataset was split in a training (70\%) and a testing (30\%) set. In addition, we further randomly split the training set in a small labeled part and a large unlabeled one, according to three different percentages for labeled objects (2\%, 5\%, 10\%). For feature representation, we used two models belonging to state-of-the-art CNN families of architectures, ResNet and DenseNet. In particular we used the smallest models offered in PyTorch library, the choice motivated by the fact that our datasets are relatively small, and so models with smaller number of parameters are expected to work better. The features were combined to generate the similarity matrix $W$, as described in Eq. \ref{eq:payoff}. The matrix for GTG model was initialized as described in the previous section. We ran the GTG algorithm up to convergence, with the pseudo-labels being computed by doing an $argmax$ over the final probability vectors.

%\input{results_figure_relative}
%%%%%%%%%%%%%%%%%%%%%%%%%%%%%%
\begin{figure*}[ht!]
\centering
\subfloat[]{\includegraphics[width = .33\textwidth]{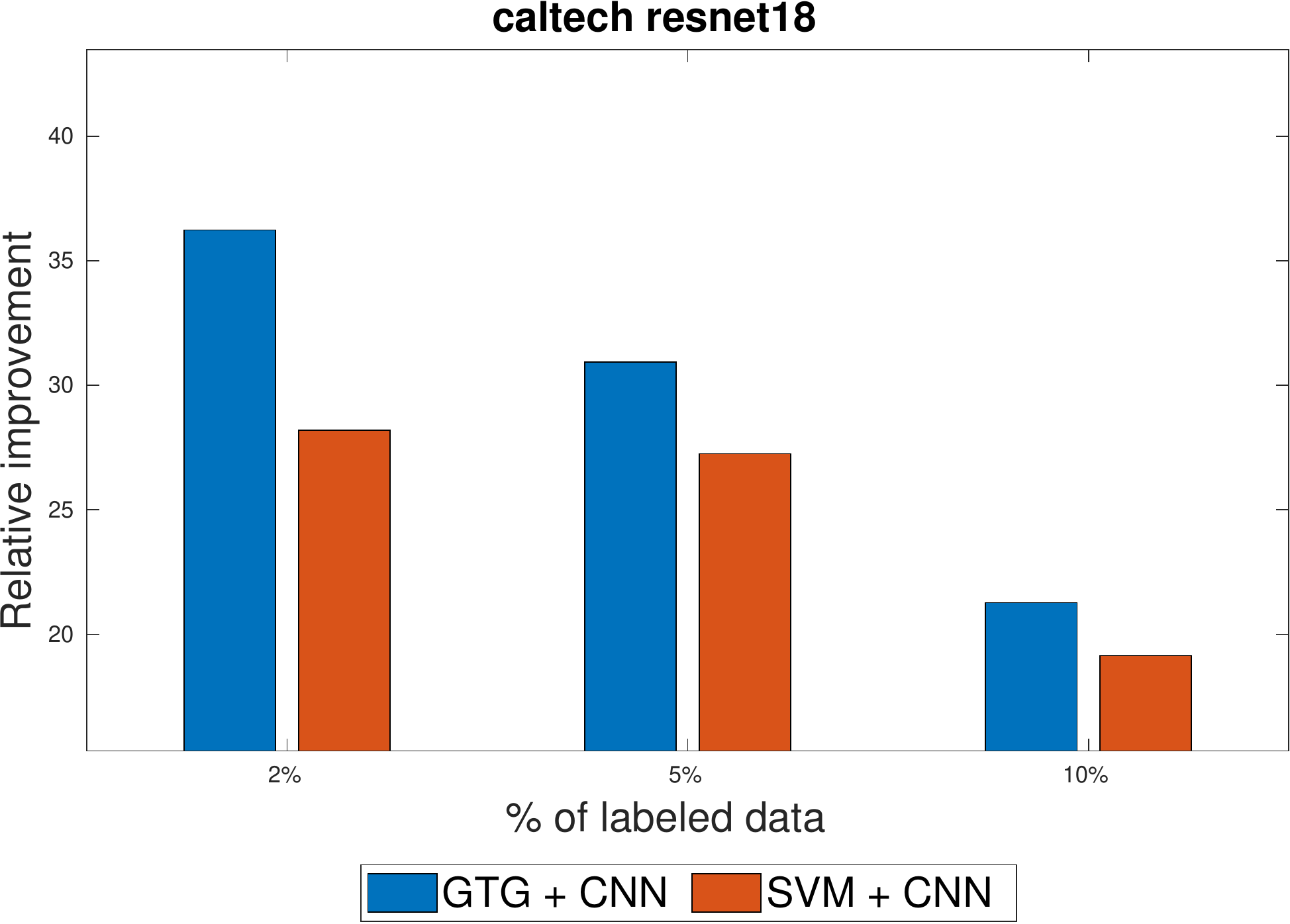}} 
\subfloat[]{\includegraphics[width = .33\textwidth]{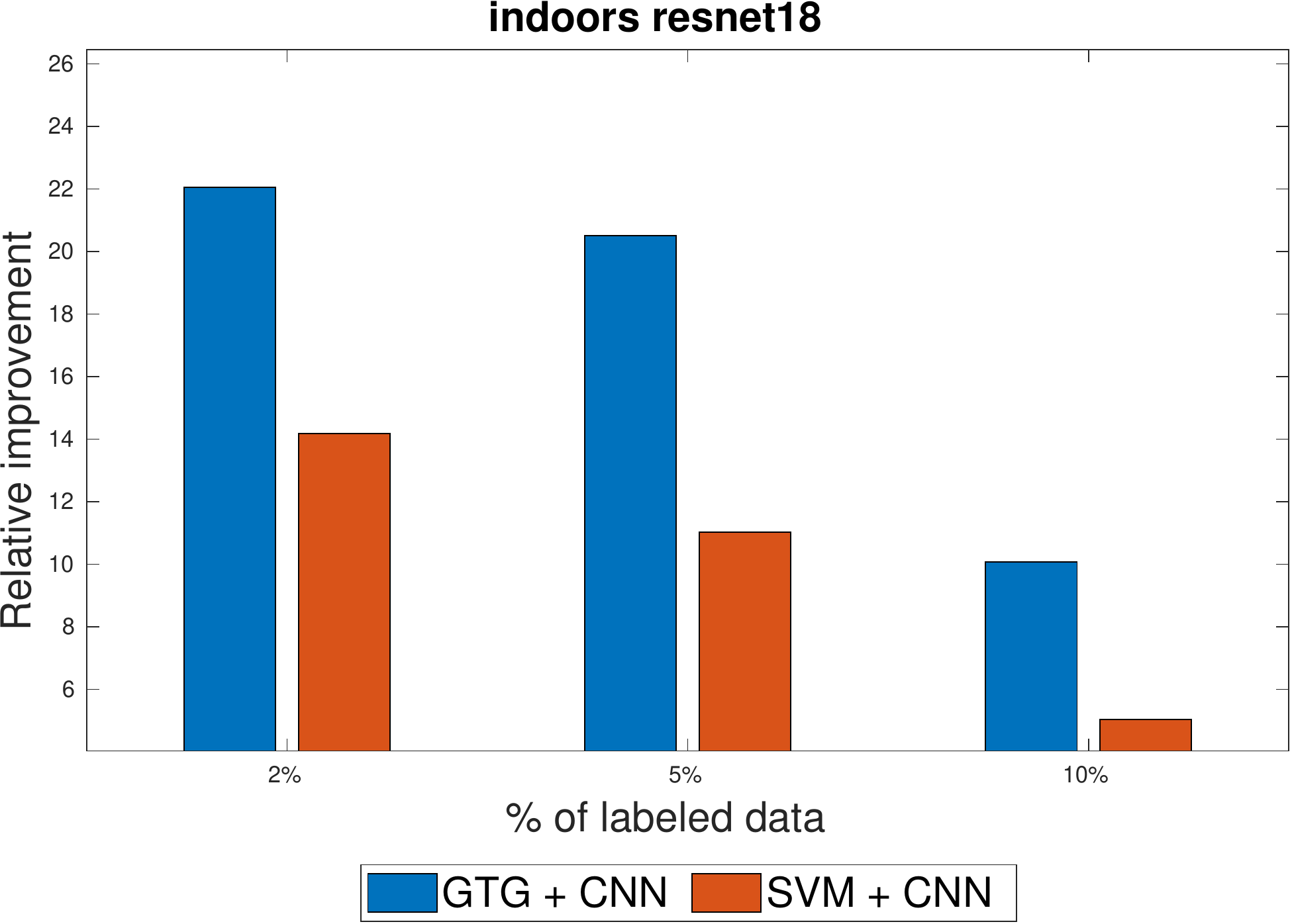}}
\subfloat[]{\includegraphics[width = .33\textwidth]{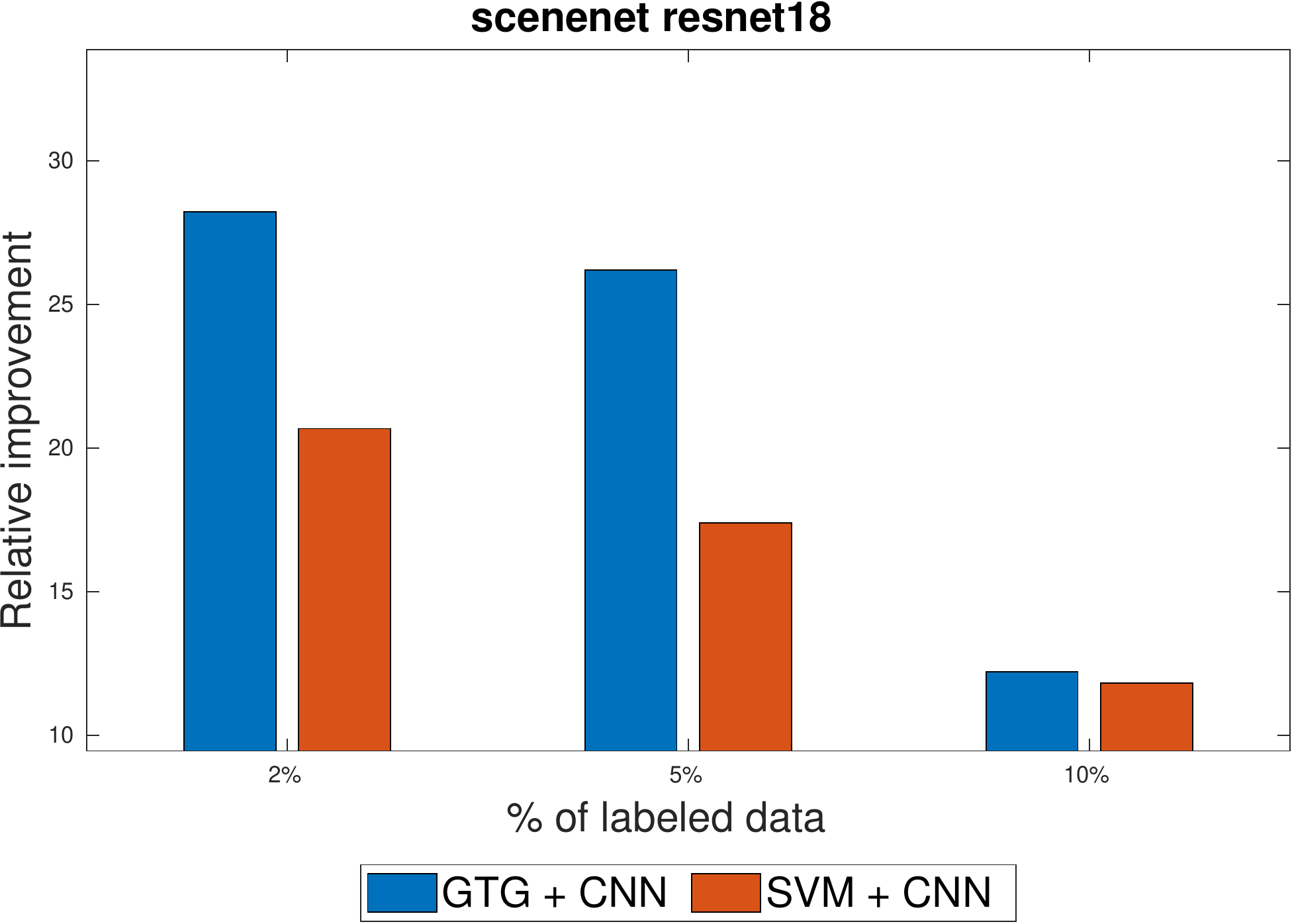}}\\
\subfloat[]{\includegraphics[width = .33\textwidth]{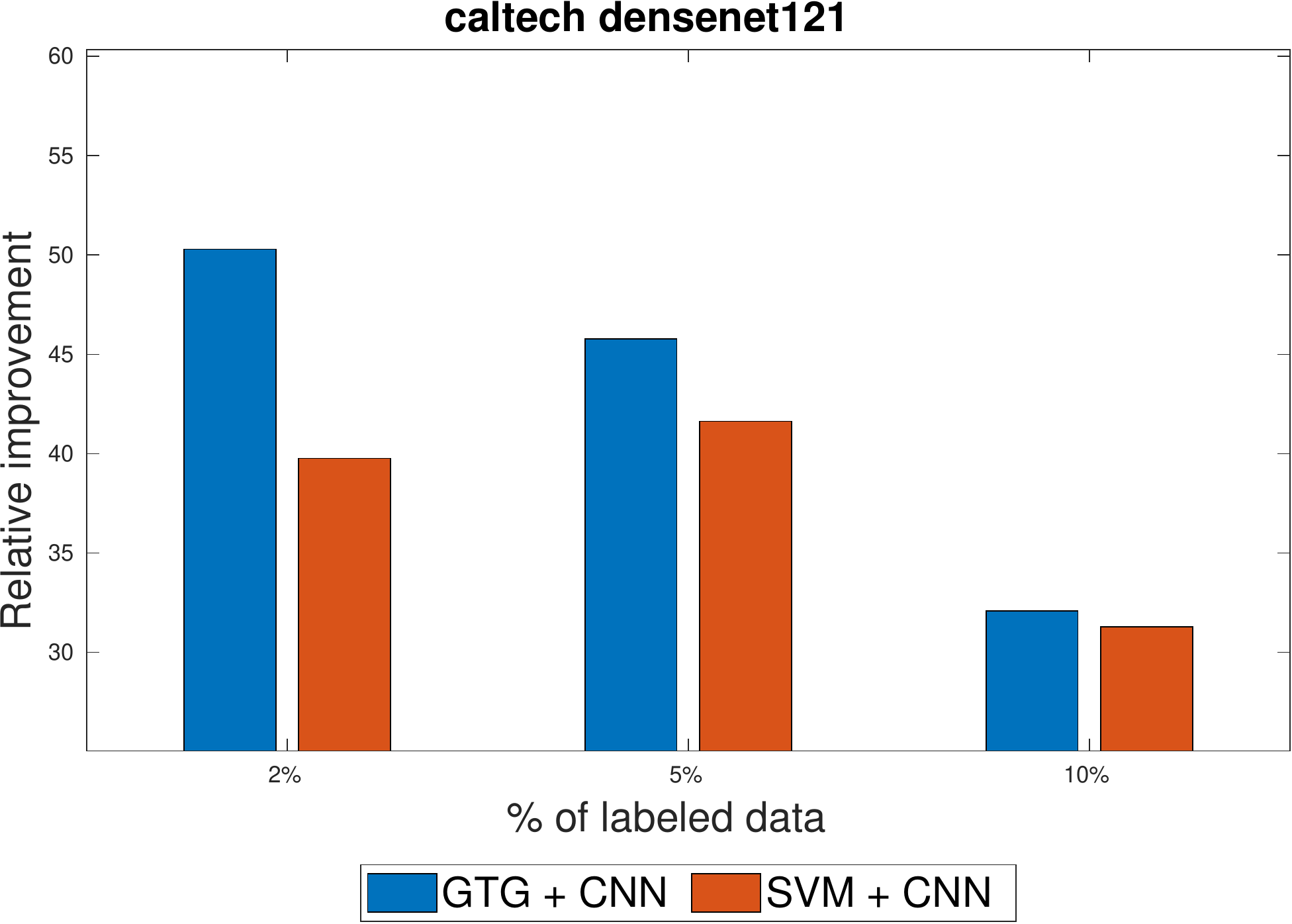}}
\subfloat[]{\includegraphics[width = .33\textwidth]{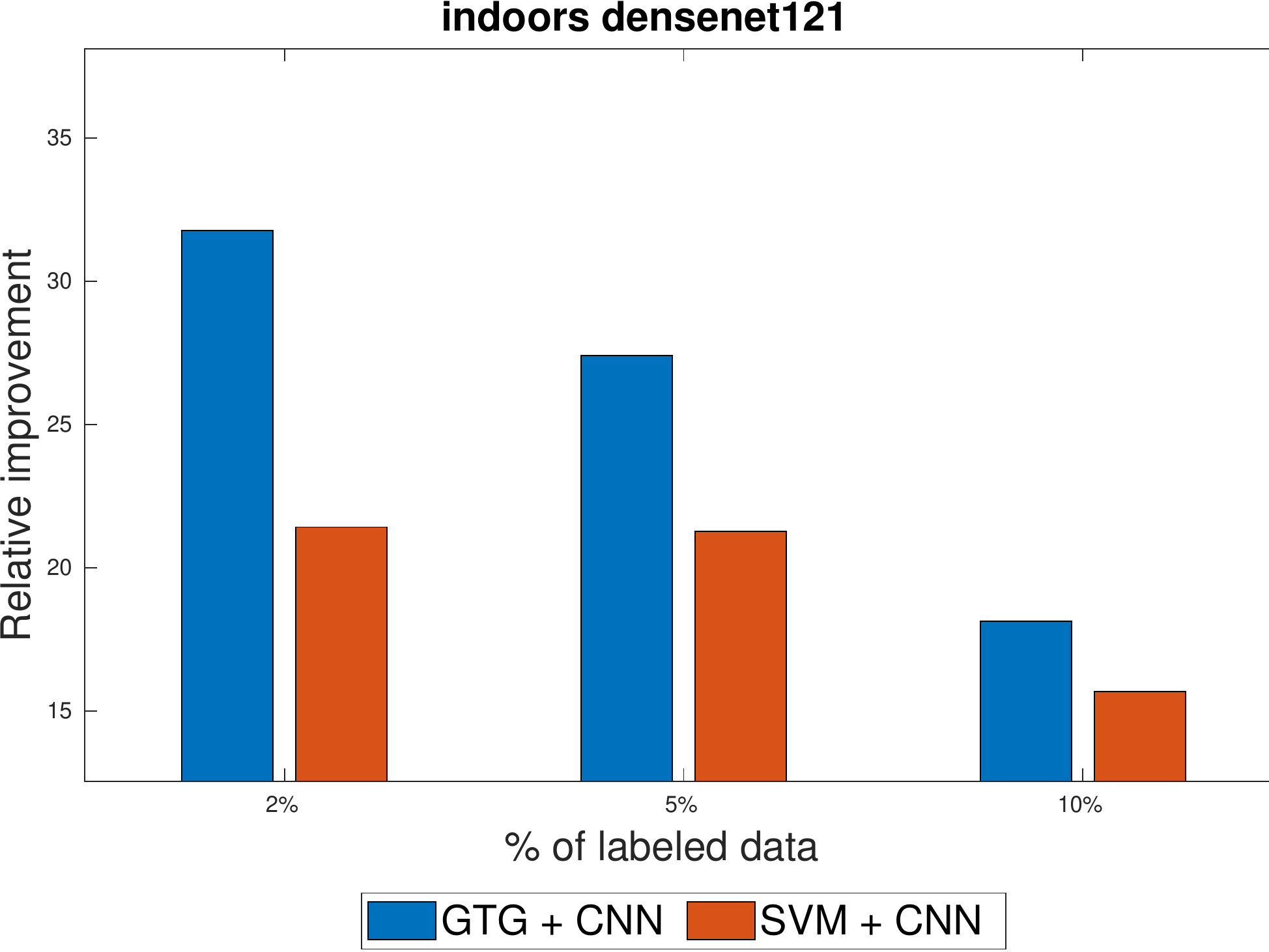}}
\subfloat[]{\includegraphics[width = .33\textwidth]{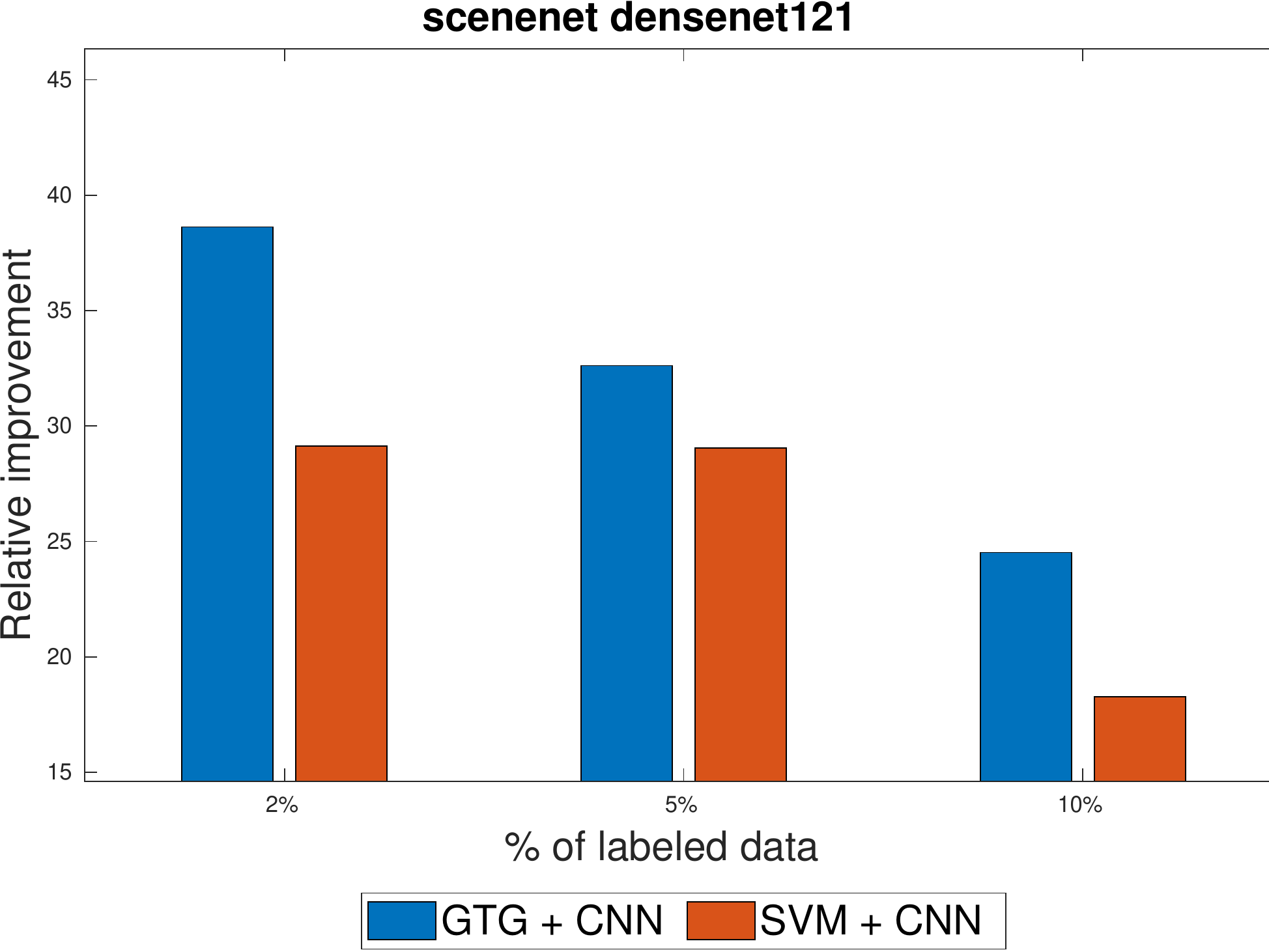}}
\caption{Results obtained on different datasets and CNNs. Here the relative improvements with respect to the CNN accuracy is reported. As can be seen, the biggest advantage of our method compared to the other two approaches, is when the number of labeled points is extremely small (2\%). When the number of labeled points increases, the difference on accuracy becomes smaller, but nevertheless our approach continues being significantly better than CNN, and in most cases, it gives better results than the alternative approach.}
\label{fig:results}
\end{figure*}
%%%%%%%%%%%%%%%%%%%%%%%%%%%%%%
We then trained \emph{ResNet18} (RN18) and \emph{DenseNet121} (DN121) in the entire dataset, by not having a distinction between labels and pseudo-labels, using Adam optimizer \cite{DBLP:journals/corr/KingmaB14} with $3 * 10^{-4}$ learning rate. We think that the results reported in this section are conservative, and can be improved with a more careful training of the networks, and by doing an exhaustive search over the space of hyper-parameters.

For comparison, we performed an alternative approach, by replacing GTG with a first-order information algorithm, namely linear SVM. While we experimented also with kernel SVM, we saw that its results are significantly worse than those of linear SVM, most likely because the features were generated from a CNN and so they are already quite good, having transformed the feature space in order to solve the classification problem linearly. No other transductive methods have been taken into consideration, since GTG has already been compared with them in \cite{DBLP:journals/neco/ErdemP12, DBLP:journals/jmiv/vascon2018}, showing that it performs better.

On Table I we give the results of the accuracy and F score on the testing set, in all three datasets, while the number of labels is only 2\% for each of the datasets ($400$ observations for Caltech-256, $200$ observations for Indoor, and $140$ observations for Scenenet). In all three datasets, and both CNNs, our results are significantly better than those of CNNs trained only in the labeled data, or the results of the alternative approach when a linear SVM is used instead of GTG. Table II and Table III give the results of the accuracy and F score while the number of labeled images is 5\%, respectively 10\%. It can be seen that with the number of labeled points increasing, the performance boost of our model becomes smaller, but our performance still gives better (or equal) results to the alternative approach in all bar three cases, and it gives significantly better results than CNN in all cases.

Figure \ref{fig:results} shows the results of our approach compared with the other approach and with the results of CNN. We plotted the relative improvement of our model and the alternative approach over CNN. When the number of labels is very small (2\%), in all three datasets we have significantly better improvements compared with the alternative approach. Increasing the number of labels to 5\% and 10\%, this trend persists. In all cases, our method gives significant improvements compared to CNN trained on only the labeled part of the dataset, with the most interesting case (only 2\% of labeled observations), our model gives 36.24\% relative improvement over CNN for \emph{ResNet18} and 50.29\% relative improvement for \emph{DenseNet121}.

\section{Conclusions and Future Work}

In this paper, we proposed and developed a game-theoretic model which can be used as a semi-supervised learning algorithm in order to label the unlabeled observations and so augment datasets. Different types of algorithms (including state-of-the-art CNNs) can then be trained on the extended dataset, where the ``pseudo-labels'' can be treated as normal labels.

Our method is not the only semi-supervised learning model used to train deep learning methods, and at this stage, we do not claim that our method is the best one. However, to the best of our knowledge, the other methods are directed towards deep learning and incorporated within the learning algorithm itself. On the contrary, we offer a different perspective, developing a model which is algorithm-agnostic, and which doesn't even need the data to be on feature-based format.

Part of the future work will consist on tailoring our model specifically towards convolutional neural networks and to make comparisons with other semi-supervised learning algorithms. In addition to this, we believe that the true potential of the model can be unleashed when the data is in some non-traditional format. In particular, we plan to use our model in the fields of bio-informatics and natural language processing, where non-conventional learning algorithms need to be developed. \SV{A direct extension of this work is to embed into the model the similarity between classes which has been proven to significantly boost the performances of learning algorithms.}

\section*{Acknowledgements}
This work was supported by Samsung Global Research Outreach Program. We thank the anonymous reviewers for their suggestions to improve the paper.

\bibliography{icpr-bib}
\bibliographystyle{IEEEtran}
\end{document}